\pdfoutput=1

\documentclass[11pt,a4paper]{article}
\usepackage[hyperref]{acl2020}
\usepackage{times}
\usepackage{latexsym}

\usepackage{microtype}

\usepackage{times}
\usepackage{latexsym}
\usepackage{multirow}
\usepackage{amsmath}

\usepackage{url}

\usepackage{graphicx}
\usepackage{stmaryrd}

\aclfinalcopy 

\title{Machine Translation with Unsupervised Length-Constraints}

\author{Jan Niehues \\
  Department of Data Science and Knowledge Engineering (DKE), Maastricht University\\
  \texttt{jan.niehues@maastrichtuniversity.nl} }

\date{}

\begin{document}
\maketitle
\begin{abstract}

We have seen significant improvements in machine translation due to the usage of deep learning. While the improvements in translation quality are impressive, the encoder-decoder architecture enables many more possibilities. In this paper, we explore one of these, the generation of constraint translation.  We focus on length constraints, which are essential if the translation should be displayed in a given format.

In this work, we propose an end-to-end approach for this task. Compared to a traditional method that first translates and then performs sentence compression, the text compression is learned completely unsupervised. By combining the idea with zero-shot multilingual machine translation, we are also able to perform unsupervised monolingual 
sentence compression. In order to fulfill the length constraints, we investigated several methods to integrate the constraints into the model.

Using the presented technique, we are able to significantly improve the translation quality under constraints. Furthermore, we are able to perform unsupervised monolingual 
sentence compression.

\end{abstract}

\section{Introduction}

Neural machine translation (NMT)~\cite{sutskever_sequence_2014,bahdanau_neural_2014} exploits neural networks to directly learn to transform sentences in a source language to sentences in a target language. 
This technique has significantly improved the quality of machine translation \cite{bojar_findings_2016,cettolo_iwslt_2015}. The advances in quality also allow for the application of this technology to new real-world applications.

While research systems often purely focus on a high translation quality, real-world applications often have additional requirements for the output of the system. One example is the mapping of markup information from the source text to the target text \cite{zenkel_adding_2019}. In this work, we will focus on another use case, the generation of translations with given length constraints. Thereby, we focus on compression. That means the target length is shorter than the actual length of the translation. When translating from one language to another, the length of the source text is usually different from the length of the target text. While for most applications of machine translation this does not pose a problem, for some applications this significantly deteriorates the user experience. For example, if the translation should be displayed in the same layout as the source text (e.g. in a website), it is advantageous if the length stays the same. Another use case are captions for videos. A human is only capable of reading text with a given speed. For an optimal user experience, it is therefore not only important to present an accurate translation, but also to present the translation with a maximum number of words. 

A first approach to address this challenge would be to use a cascade of a machine translation and sentence compression system. In this case, we would need training data to train the machine translation system and additional training data to train the sentence compression system. It is very difficult and sometimes even impossible to collect the training data for the sentence compression task. Furthermore, we need a sentence compression model with a parametric length reduction ratio. For a supervised model, we would therefore need examples with different length reduction ratios. Therefore, this work focuses on unsupervised sentence compression. In this case, the cascaded approach cannot directly be applied and we will start with an end-to-end approach to length-constraint translation.

While our work focuses on the end-to-end approach to translation combined with sentence compression, monolingual sentence compression is another important task. For example, human-generated c  aptions are often not an accurate transcription of the audio, but in addition the text is shortened. This is due to cognitive processing constraints. The user is able to listen to more words in a given time than he can read in the same amount of time. When combining the length-constrained machine translation with the idea of zero-shot machine translation,
the proposed method is also able to perform monolingual sentence compression. Compared to related work, this methods will learn the text compression in an unsupervised manner without ever seeing a compressed sentence.

The main contribution of this work is an end-to-end approach to length-constraint translation by jointly performing machine translation and sentence compression. We are able to show that for this task an end-to-end approach outperforms the cascade of machine translation and unsupervised sentence compression. 

Secondly, we perform an in depth analysis how to integrate additional constraints into neural machine translation. We investigate methods to restrict the search to translations with a given length as well as adapting the transformer-based encoder-decoder. In the analysis, we also investigated the portability of the technique to encode other constraints. Therefore, we applied the same techniques to avoiding difficult words.

A third contribution of this work is to extend the presented approach to unsupervised monolingual sentence compression. By combining the presented approach with multilingual machine translation, we are able to also generate paraphrases with a given length constraint. The investigation shows that 
a system that is trained on several languages is able to successful generate monolingual paraphrases.

\section{Constraint decoding}

In this work, we address the challenge of generating machine translations with additional constraints. The main application is length-constrained translation. That means that we want to generate a translation with a given target length. We focus thereby on the case of shortening the translations. While the length can be measured in words, sub-word tokens or letters, in the experiments we measured the length by sub-word tokens. 

A first straightforward approach is to restrict the search space to generate only translations with a given length. The length of the output is modeled by the probability of the end-of-sentence (EOS) token. By modifying this probability, we introduce a hard constraint that is always fulfill.

Based on the experiments with this type of length constraints, we then investigate methods to include the length constraints directly into the model. For that we used two techniques successfully used in encoder-decoder models: embeddings and positional encodings. In this case, the target length is modeled as a soft constraint. However, by combining both techniques, we can again achieve a hard constraint.

\subsection{Search space}

A first strategy to include the additional constraints is to ignore them during training and restrict the search space during inference to translations that fulfill the constraint. For length constraints, this can be achieved by manipulating the end-of-sentence token probability. First, we need to ensure that the EOS token is not generated before the desired length of output $J$. This can be ensure by setting the probability for the end-of-sentence token to zero for all positions before the desired length and renormalizing the probability.

    \begin{align}
    p'(y_j|x_1,\ldots,x_I,y1,\ldots,y_{j-1})= \\ \nonumber
    \left\{ 
    \begin{array}{cc}
    \frac{p (y_j|x_1,\ldots,x_I,y1,\ldots,y_{j-1})}{1-p (EOS|x_1,\ldots,x_I,y1,\ldots,y_{j-1})} & y_j \ne EOS \\
    0 & y_j = EOS
    \end{array}
    \right.
    \end{align}

Finally, we ensure to stop the search at the desired length by setting the probability of the end-of-sentence token to one if the output sequence has reached this length.

    \begin{align}
    p'(y_j|x_1,\ldots,x_I,y1,\ldots,y_{j-1})= \\ \nonumber
    \left\{ 
    \begin{array}{cc}
    0 & y_j \ne EOS \\
    1 & y_j = EOS
    \end{array}
    \right.
    \end{align}

While this approach will guarantee that the output of the translation systems always meets the length condition (hard constraint), it has also one major drawback. Until the system reaches the constraint length, the system is not aware of how many words it is still allowed to generate. Therefore, it is not able to shorten the beginning of the sentence in order to fulfil the length constraint. 

Motivated by this observation, we investigate methods to integrate the length constraint into the model and not only apply it during inference.

\subsection{Pseudo-supervised training}

The first challenge we need to address when including the length constraint into the model itself is the question of training data. While there is large amounts of parallel training data, it is hard to acquire training data with length constraints. Therefore, we investigate methods to train the model with standard parallel training data.

We perform the training by a type of pseudo-supervision. For each source sentence, in training, we also know the translation and thereby its length. The main idea is that we now assume this sentence was generated with the constraint to generate a translation with exactly the length of the given translation. 
Of course, this is mostly not the case. The human translator generated a translation that appropriately expresses the meaning of the source sentence and not a sentence that fulfills the length constraints. 

Therefore, learning in this case is more difficult. Since the translations are generated without the given length constraints the systems might learn to simply ignore the length information and instead generate a normal translation putting all the information of the source sentence into the target sentence. In this case, we would not have the possibility to control the target length by specifying our desired length. 

Therefore, we continue to investigate different possibilities how to encode the constrained target length in the model architecture.

\subsection{Length representation}

In this work, we investigate three different methods to represent the target length in the model. In all cases our training data consists of a source sentence $X=x_1,\ldots,x_I$, a target sentence $Y=y_1,\ldots,y_J$ and the target length $J$.

\paragraph{Source embedding}
A first method is to include the target length into the source sentence as an additional token. This is motivated by successful approaches for multilingual machine translation \cite{ha_toward_2016}, domain adaptation \cite{kobus_domain_2017} and formality levels \cite{sennrich_controlling_2016}. We change the training procedure to not use $X$ as the input to the encoder of the NMT system, but instead $J,X$. Thereby, the encoder will learn a embedding for each target length seen during training. 

There are two challenges using this approach. First, the dependency between the described length $J$ and the output $Y$ is quite long within the model. Therefore, the model might ignore the information and just learn to generate the best translation for a given source sentence. Secondly, the representations for all possible target lengths are independent from each other. The embedding for length $20$ is not constrained to be similar to the one of length $21$. This poses a special challenge for long sentences which occur less frequently and therefore the embedding of these lengths will not be learned as well as the frequent ones.

\paragraph{Target embedding}

We address the first challenge by integrating the length constraint directly into the decoder. This is motivated by similar approaches to supervised sentence compression \cite{kikuchi_controlling_2016} and zero-shot machine translation \cite{ha_effective_2017}. We incorporate the information of the number of remaining target words at each target position. For one, this should ensure that the length information is not lost during the decoding process. Secondly, by embedding smaller numbers which occur more frequently in the corpus towards the end of the sentence, the problem of rare sentence lengths does not matter that much. 

Formally, at each decoder step $j$ the baseline model starts with the word embedding of the last target word $y_{j-1}$. In the original transformer architecture, the positional encoding is applied on top of the embedding to generate the first hidden representation.
\begin{equation}
    h_0 = pos(emb(y_{j-1}),j)
\end{equation}
In our proposed architecture, we include the number of remaining target words to be generated $J-j$. 
We concatenate $h_0$ with the length embedding and then apply a linear translation and a non-linearity to reduce the hidden size to the one of the original word embedding

\begin{equation}
    h'_0 = relu(lin(cat(h_0,lenEmb(J-j)))
\end{equation}

The proposed architecture allows the model to consider the number of remaining target words at each decoding step. While the baseline model will only cut the end of the sentences, the model is able to shorten already consider the constraints at the beginning of the sentence.

\paragraph{Positional encoding}

Finally, we also address the challenge of representing sentence lengths that are less frequent. The transformer architecture \cite{vaswani_attention_2017} introduced the positional encoding. This encodes the position within the sentence using a set of trigonometric functions. While their method encodes the position relative to the start of the sentence, we  follow \cite{takase_positional_2019} to encode the position relative to the end of the sentence. Thereby, at each position we encode the number of remaining words of the sentence. 
Formally, we replace $h_0 = pos(emb(y_{j-1}),j)$ by $h*_0 = pos(emb(y_{j-1}),J-j)$.

\subsection{Additional constraints}
\label{addConstraints}
Besides constraining the number of words, other constraints can be implemented as easily using the same framework. In this work, we show this by limiting the number of complex and difficult words. One use case is the generation of paraphrases in simplified language. A metric to measure text difficulty, the Dale-Chall Readability metric \cite{chall_readability_1995}, for example, counts such difficult words. In an NMT system, longer words are typically split into subword units by Byte Pair encoding \cite{sennrich_neural_2016}. A complex word like \textit{marshmallow} is split into several parts like \textit{mar@@ shm@@ allow}. Thereby, \textit{@@} indicates that the word is not yet finished.

We can encourage the system to use less complex words that need to be represented by several BPE tokens by counting the number of tokens that do not end a word. In the proposed encoding scheme these are all tokens that end in \textit{@@}. During decoding, we then try to generate translations with a minimal number of these tokens.

\section{Evaluation}

The lack of appropriate data is not only a challenge for training the model but also for evaluating. The default approach to evaluate a machine translation methods is to compare the translation by the system with human translation using some automatic metric (e.g. BLEU \cite{papineni_bleu:_2001}). 

In our case, we would need to have a human-generated translation, which also fulfills the additional constraints. For example, translation with a length that is shorten to 80\% of the input. 

Since this type of translation data is not available, we investigate methods to compare the length-constraint output of the system with standard human translation that do not fulfill any specific constraints.

\subsection{Word matching metrics}

While there is significant research in automatic metrics for machine translation \cite{ma_results_2018,ma_results_2019}, BLEU is still the most commonly used metric. Therefore, a first approach would be to use BLEU to compare the automatic translation with length constraints with the human translation without constraints. If we were using length constraints, this would lead to low BLEU score due to the length penalty of the metric. But since all systems must fulfill the length constraint, the penalty would be the same for all output and we could still compare between the different outputs.

\begin{table*}
\begin{tabular}{ll}
Reference: & So CEOs, a little bit better than average,  but here's where it gets interesting. \\
Baseline: & CEOs are a little bit \\
Constraint: & the CEOs are interesting .\\
\end{tabular}
\caption{\label{table:example1} Example of constraint translation }
\end{table*}

A problem of using the BLEU metric for this task is illustrated by the example translations in Table \ref{table:example1}. The Baseline system only uses the length constraint for restricting the search space. In the constraint system, we are using the length constraint also as additional embeddings in the decoder. Looking at this example sentence, a human would always rate the constraint translation better as the baseline translation. The problem of the latter model is that it is often generating a prefix of the full translation. While this does not lead to a good constrained translation, it still leads to a relative high BLEU score. In this case, we have one matching 3-gram, two bigrams and four unigrams. 

In contrast, the constraint model only contains words matching the reference scattered over the sentence. Therefore, in this case, we only have two unigram matches. Guided by this observation, we used different metrics to evaluate the models.

\subsection{Embedding-based metrics}

In order to address the challenges mentioned in the last section, we used metrics that are based on sentence embeddings instead of word or character-based representation of the sentence. This way it is no longer important that the words occur in the same sequence in automatic translation and reference. Based on the performance of the automatic metrics in the WMT Evaluation campaign in 2018, we used RUSE \cite{shimanaka_ruse:_2019} metric. It uses sentence embeddings from three different models: InferSent, Quit-Thought and Universal Sentence Encoder. Then the quality is estimated by an MLP based on the representation of the hypothesis and the reference translation.

\section{Experiments}

\subsection{Data}

We train our systems on the TED data from the IWSLT 2017 multilingual evaluation campaign \cite{cettolo_overview_2017}. The data contains parallel data between German, English, Italian, Dutch and Romanian. We create three different systems. The first system is only trained on the German-English data, the second one is trained on German-English and English-German data and the last one is trained on \{German, Dutch, Italian, Romanian\} and English data in the both directions.

The data is preprocessed using standard MT procedures including tokenization, truecasing and byte-pair encoding with 40K codes. For model selection, the checkpoints performing best on the validation data (dev2010 and tst2010 combined) are averaged, which is then used to translate the tst2017 test set.

In the experiments, we address two different targeted lengths. In order to not use any information from the reference, we measure length limits relative to the source sentence length by counting subword units. We aim to shorten the translation to produce output that is 80\% and 50\% of the source sentence length. In all experiments, the length is measured by the number of sub-word tokens.

\subsection{System}

We use the transformer architecture \cite{vaswani_attention_2017} and increase the number of layers to eight. The layer size is 512 and the inner size is 2048. Furthermore, we apply word dropout with $p=0.1$ \cite{gal_theoretically_2016}. In addition, layer dropout is used with $p=0.2$ as in the original work. We use the same learning rate schedule as in the original work. The implementation is available at \textit{https://github.com/quanpn90/NMTGMinor.git}.

\subsection{Task difficulty}
Initially, we wanted to investigate the difficulty of having the additional length constraints. Therefore, we used the length of the human reference translation as a first target length. One could even argue that should make the typical machine translation easier, since some information about the translation is known. The results of this experiment are shown in Table \ref{taskDifficulty}. Since we do not perform compression in this experiment, the aforementioned problem with BLEU should not apply here.

\begin{table}[ht]
 \begin{center}
   \begin{tabular}{|l|c|c|} \hline 
Model & BLEU & RUSE \\ \hline
Baseline & 30.80 & -0.085\\
Only Search & 28.32 & -0.124\\
Source Emb & 28.56 & -0.126 \\
Decoder Emb & 27.88 & -0.140\\
Decoder Pos & 28.80 & -0.138\\ \hline
\end{tabular}   
\end{center}
\caption{\label{taskDifficulty} Using oracle length}
\end{table}

However, the results indicate the baseline system achieves the best BLEU score as well as the best RUSE score. All other models generate translations that perfectly fit the desired target length, but this leads to a drop in translation quality. Therefore, even if the target length is the same as the one of the reference translation, by  restriction increases the difficulty of the problem. One reason could be that the machine translation system rarely generates translations which exactly match the reference. By forcing the translation to have an exact predefined length, we are increasing the difficulty of the problem.

\subsection{Length representation}

In a first series of experiments, we analyzed the different techniques to encode the length of the output. First, we are interested in whether the different length representations are able to enforce an output that has the length we are aiming at (soft constraints). For the German to English translation task, the length of the different encoding versions are shown in Table \ref{lengthDEEN}. We define the length as the average difference between the targeted output given in BPE units and the output of the translation system.

First, without adding any constraints, the models generate translations that differ by 3.9 and 10.29 words from the targeted length. By specifying the length in the source side, we can reduce the length difference to half a word in the case of a targeted length of 80\% and one and a half words in the case of 50\% of the source length. The models using the decoder embeddings and the decoder positional encoding where able to nearly perfectly generate translation with the correct number of words.

\begin{table}[ht]
 \begin{center}
   \begin{tabular}{|l|c|c|} \hline 
Encoding & \multicolumn{2}{c|}{Avg. length difference} \\
& 80\% & 50\% \\ \hline
Baseline & 3.90 & 10.29\\
Source Emb & 0.55 & 1.40\\
Decoder Emb & 0.07 & 0.16\\
Decoder Pos & 0.09 & 0.19\\ \hline
\end{tabular}   
\end{center}
\caption{\label{lengthDEEN} Avg. Length distance }
\end{table}

Besides fulfilling the length constraints, the translations needs to be accurate. Since we wanted to have a fair comparison, we evaluated the output when using a restrict search space, so that only translation with the correct number of words are generated (hard constraints). The results are summarized in Table \ref{MTQdeen}

\begin{table}[ht]
 \begin{center}
   \begin{tabular}{|l|c|c|} \hline 
Encoding & \multicolumn{2}{c|}{RUSE} \\
& 80\% & 50\% \\ \hline
Baseline & -0.272 & -0.605\\
Source Emb & -0.263 & -0.587 \\
Decoder Emb & -0.2469 & -0.555\\
Decoder Pos & -0.2598 & -0.577\\ \hline
\end{tabular}   
\end{center}
\caption{\label{MTQdeen} German-English translation quality }
\end{table}

As shown be the results, we see improvements in translation quality when using the source embedding within the encoder. We have further improvements if we represent the targeted length within the decoder. In this case, we can improve the RUSE score by 2\% and 5\% absolute. The decoder encodings perform similarly, with small advantage for using embeddings and not positional encodings.Therefore, in the remaining of the experiments we use the embeddings.

\subsection{Multi-lingual}

In a second series of experiments, we combine the constraint translation approach with multi-lingual machine translation. The combination of both offers the unique opportunity to perform unsupervised sentence compression. We can treat the translation of English to English as a zero-shot direction \cite{johnson_googles_2017,ha_toward_2016}. This has not been addressed in traditional multi-lingual machine translation, since in this case the model will often just copy the source sentence to the target one. By adding the length constraints, we force the mode to reformulate the sentence in order to fulfil the length constraint.

The results for these experiments are shown in Table \ref{MultilingualMT}. In this case, we compared three scenarios. First, a model trained only to translation from German to English. Secondly, a model trained to translate from German to English and English to German. Finally, a model trained on 4 language to and from English.

\begin{table*}[ht]
 \begin{center}
   \begin{tabular}{|l|c|c|c|c|c|c|c|c|} \hline 
   & \multicolumn{4}{c|}{Target Length 0.8} & \multicolumn{4}{c|}{Target Length 0.5}\\
 &  \multicolumn{2}{c|}{Baseline} & \multicolumn{2}{c|}{Dec. Emb} &  \multicolumn{2}{c|}{Baseline} & \multicolumn{2}{c|}{Dec. Emb} \\
Model & DE-EN & EN-EN & DE-EN & EN-EN & DE-EN & EN-EN & DE-EN & EN-EN \\ \hline
DE-EN & -0.272 & & -0.247 & & -0.587 & & -0.554 &\\
DE+EN & -0.264 & -0.817 & -0.223& -0.905 & -0.598 & -0.954 & -0.523 & -0.978\\
All & -0.225 & -0.102 & -0.214 & 0.020 & -0.560 & -0.525 & -0.548 & -0.481\\ \hline
\end{tabular}   
\end{center}
\caption{\label{MultilingualMT} Multi-lingual systems}
\end{table*}


First of all, since the models are trained on relative small data, we always gain when using more language pairs. Secondly, for all models training from German to English, the decoder embedings is clearly better than the baseline. Finally, to perform paraphrasing, we need a multi-lingual system with several language pairs. Both model trained only on the German to English and English to German data fail to generate adequate translation. In contrast, if we look at the translation from English to English for the multilingual model, the scores are clearly better than the ones from German to English. Furthermore, again, the system with decoder embeddings is clearly better than the baseline system.

In addition, we performed the same experiment with a target length of half the source length (Table \ref{MultilingualMT}). Although the absolute scores are significant lower since the model has to reduce the length further, the tendency is the same for this direction.

\subsection{End2End vs. Cascaded}

\begin{table}[ht]
 \begin{center}
   \begin{tabular}{|l|l| c|c|c|c|} \hline 
Length &  Model & DE-EN & EN-EN \\ \hline
\multirow{3}{*}{0.8} & End2End & -0.247 & 0.020 \\
 & Cascade & -0.259 & -0.118 \\
 & Cascade Fix. Pivot & & -0.166 \\ \hline
\multirow{3}{*}{0.5} & End2End & -0.555 & -0.481\\
 & Cascade & -0.575 & -0.521\\
 & Cascade Fix. Pivot & & -0.544\\ \hline
\end{tabular}   
\end{center}
\caption{\label{E2E} Comparison of End-to-End and Cascaded approach }
\end{table}

In this work, we are able to combine machine translation and sentence compression. In a third series of experiments, we wanted to investigate the advantage of modelling it in an end-to-end fashion compared to a cascade of different models. We performed this investigation again for two tasks: German to English and English to English.

The cascade system for German to English, first translates the German text to English with a baseline machine translation system. In a second step, the output is compressed with the multi-lingual MT system. For the English-to-English system, the cascade system removes the zero-shot condition. Therefore, we first translate from English to German with the baseline system and then translate with length contrasted from German to English. In \textit{cascade fix pivot} also the English to German system already fulfill the length constraint.

As shown in Table \ref{E2E}, in all condition, the end-to-end approach outperforms the cascaded version. This is especially the case for the English-to-English machine translation. Compared to multi-lingual machine translation, for this tasks it seems to be beneficial to perform the zero-shot tasks instead of using a pivot language.

\subsection{Simplification}
 \begin{table*}[ht]
 \begin{center}
   \begin{tabular}{|l|r|r|r|r|r|r|} \hline 
Metric &  \multicolumn{2}{c|}{DE-EN} & \multicolumn{2}{c|}{DE+EN} & \multicolumn{2}{c|}{All}\\
& Base & Simp. & Base & Simp. & Base & Simp \\ \hline
BPE tokens & 1961 & 1053 & 1978 & 1041 & 1899 & 991 \\
DCI & 7.63 & 7.47 & 7.69 & 7.5 & 7.66 & 7.45 \\
FRE & 83.86 & 86.18 & 84.31 & 85.49 & 82.98 & 85.59 \\
BLEU & 30.80 & 30.62 & 32.25 & 31.38 & 32.84 & 31.29 \\
RUSE & -0.085 & -0.092 & -0.082 & -0.080 & -0.042 & -0.084 \\ \hline
\end{tabular}   
\end{center}
\caption{\label{Simplification} Simplification }
\end{table*}
In a last series of experiments (Table \ref{Simplification}), we investigate the ability of the framework to generate simpler sentence. As described in Section \ref{addConstraints}, we concentrate on reducing the number of rare and complex words. 
Again, we are using the decoder embedding to represent the amount of BPE units in the sentences. We use a system for 1 language pair, 2 language pairs and the system using 8 language pairs. First, the system is able to reduce the number of BPE tokens in the text significantly. The amount of tokens is reduce by up to 48\%. Since the number of tokens is nearly keep the same, this is also reflected in a better readability. On the other hand, we see that the translation quality is only affected slightly.

\subsection{Qualitative Results}

For the length restricted system, we also present examples in Table \ref{Examples}. The translation were generated with the multi-lingual system using restricted search space with 0.8 times and 0.5 times the length of the source length. The length is thereby measured using the number of subword tokens.

\begin{table*}[h]
 \begin{center}
   \begin{tabular}{|l|l|} \hline 

Source: & Und, obwohl es wirklich einfach scheint, ist es tatsächlich richtig schwer, \\ & weil es Leute   drängt sehr schnell zusammenzuarbeiten. \\
Reference: & And, though it seems really simple, it's actually pretty hard  because it \\ & forces people  to   collaborate very quickly. \\
Base 0.8: & and even though it really seems simple , it is actually really hard , because \\ & it really pushes \\
Dec. Emb. 0.8 : & and although it really seems simple , it is really hard because it drives \\ &  people to  work together . \\
Base 0.5 : & and even though it really seems simple , it is really hard \\
Dec. Emb. 0. 5: & it is really hard because it drives people to work together . \\ \hline

Source: & Konstrukteure erkennen diese Art der Zusammenarbeit als Kern eines \\ & iterativen Vorgangs. \\
Reference: & Designers recognize this type of collaboration  as the essence of the \\ & iterative process. \\
Base 0.8: &  now , traditional constructors recognize this kind of collaboration as the core \\
Dec. Emb. 0.8 & designers recognize this kind of collaboration as the core of iterative . \\

Base 0.5: & now , traditional constructors recognize this kind \\ 
Dec. Emb: 0.5 & developers recognize this kind of collaboration . \\ \hline



\end{tabular}   
\end{center}
\caption{\label{Examples} Examples }
\end{table*}

In the examples we see clearly the problem of the baseline model when using a restrict search space. The model mainly outputs the prefix of the long translation and do not try to put the main content into the shorter segment. In contrast, the system using the decoder embeddings are aware when generating a word how much space the still have to fill the content. Therefore, they do not just cut part of the sentence, but compress the sentence and extract the most important part of the sentence. While the first example is more concentrating on the second part of the original sentence, the second one is focusing at the beginning. Although the model reducing the length by 50\% have to remove some content of the original sentence, the sentence is still understandable.

\section{Related Work}

The most common approach to model the target length within NMT is the use of coverage models \cite{tu_modeling_2016}. More recently, \cite{lakew_controlling_2019} used similar techniques to generate translation with the same length as the source sentence. Compared to these works, we tried to significantly reduce the length of the sentence and thereby have the situation where the training and testing condition differ significant. 
This work on length-controlled machine translation is strongly related to sentence compression, where the compression is performed in the monolingual case. First approach used rule-based approaches \cite{dorr_hedge_2003} for extractive sentence compression. In abstractive compression methods using syntactic translation \cite{cohn_sentence_2008} and phrase-based machine translation were investigated \cite{wubben_sentence_2012}. The success of encoder-decoder models in mainly areas of natural language processing \cite{sutskever_sequence_2014,bahdanau_neural_2014} motivated it success full application to sentence compression. \cite{kikuchi_controlling_2016} and \cite{takase_positional_2019} investigated an approach to directly control the output length. Although their methods uses similar techniques to ours, the model is trained in a supervised way. Motivated by recent success in unsupervised machine translation \cite{artetxe_unsupervised_2018,lample_unsupervised_2018}, a first approach to learn text compression in an unsupervised fashion was presented in \cite{fevry_unsupervised_2018}. Text compression in a supervised fashion for subtitles was investigated in \cite{angerbauer_automatic_2019}.

In contrast to text compression, the combination of readability and machine translation has been researched recently. \cite{agrawal_controlling_2019} presented an approach to model the readability using source side annotation. In contrast to our work, they concentrated on the scenario where manually created training data is available. In \cite{marchisio_controlling_2019} the authors specified the desired readability difficulty either by a source token or though the architecture by different encoder. While they concentrate on a single task and have only a limited number of difficulty classes, the work presented here is able to handle a huge number of possible output class (e.g. in text compression the number of words) and can be applied for different task.

\section{Conclusion}

In this work, we present a first approach to length-restricted machine translation. In contrast to work on monolingual sentence compression, we focus thereby on unsupervised methods. By combining the results with multi-lingual machine translation, we are also able to perform monolingual unsupervised sentence compression.

We have shown that is important to include the length constraints to the decoder to achieve translations fulfilling the constraints. Furthermore, modeling the task in an end-to-end fashion improves over splitting the task into different sub-tasks. This is even true for zero-shot conditions.

\bibliography{zotero}
\bibliographystyle{acl_natbib}

\appendix

\end{document}